\documentclass[conference]{IEEEtran}
\IEEEoverridecommandlockouts
\usepackage{cite}
\usepackage{amsmath,amssymb,amsfonts}
\usepackage{algorithmic}
\usepackage{graphicx}
\usepackage{textcomp}
\usepackage[table]{xcolor}
\usepackage{multirow}
\usepackage{array}
\usepackage{tikz}
\usepackage{placeins}
\usepackage{url}
\usepackage{comment}
\usepackage[hidelinks]{hyperref}
\newcolumntype{C}[1]{>{\centering\arraybackslash}m{#1}}
\def\BibTeX{{\rm B\kern-.05em{\sc i\kern-.025em b}\kern-.08em
    T\kern-.1667em\lower.7ex\hbox{E}\kern-.125emX}}

\newcommand{\linebreakand}{%
  \end{@IEEEauthorhalign}
  \hfill\mbox{}\par
  \mbox{}\hfill\begin{@IEEEauthorhalign}
  }

\begin{document}

\title{Large Language Model for Qualitative Research: \\ A Systematic Mapping Study\\
}

\author{\IEEEauthorblockN{Cauã Ferreira Barros}
\IEEEauthorblockA{\textit{Informatics Institute} \\
\textit{Federal University of Goiás}\\
Goiânia, Brazil \\
cauabarros@ufg.br}
\and
\IEEEauthorblockN{Bruna Borges Azevedo}
\IEEEauthorblockA{\textit{Informatics Institute} \\
\textit{Federal University of Goiás}\\
Goiânia, Brazil \\
brunabazevedo@discente.ufg.br}
\and
\IEEEauthorblockN{Valdemar Vicente Graciano Neto}
\IEEEauthorblockA{\textit{Informatics Institute} \\
\textit{Federal University of Goiás}\\
Goiânia, Brazil \\
valdemarneto@ufg.br}
\and
\IEEEauthorblockN{Mohamad Kassab}
\IEEEauthorblockA{\textit{Boston University}\\
Boston, USA \\
mkassab@bu.edu}
\and
\IEEEauthorblockN{Marcos Kalinowski}
\IEEEauthorblockA{\textit{Pontifical Catholic University of Rio de Janeiro} \\
Rio de Janeiro, Brazil \\
kalinowski@inf.puc-rio.br}
\and
\IEEEauthorblockN{Hugo Alexandre D. do Nascimento}
\IEEEauthorblockA{\textit{Informatics Institute} \\
\textit{Federal University of Goiás}\\
Goiânia, Brazil \\
hadn@inf.ufg.br}
\and
\IEEEauthorblockN{Michelle C.G.S.P. Bandeira}
\IEEEauthorblockA{\textit{Faculty of Science and Technology} \\
\textit{Federal University of Goiás}\\
Aparecida de Goiânia, Brazil \\
michelle.galvao@ufg.br}
}

\maketitle

\begin{abstract}
The exponential growth of text-based data in domains such as healthcare, education, and social sciences has outpaced the capacity of traditional qualitative analysis methods, which are time-intensive and prone to subjectivity. Large Language Models (LLMs), powered by advanced generative AI, have emerged as transformative tools capable of automating and enhancing qualitative analysis. This study systematically maps the literature on the use of LLMs for qualitative research, exploring their application contexts, configurations, methodologies, and evaluation metrics. Findings reveal that LLMs are utilized across diverse fields, demonstrating the potential to automate processes traditionally requiring extensive human input. However, challenges such as reliance on prompt engineering, occasional inaccuracies, and contextual limitations remain significant barriers. This research highlights opportunities for integrating LLMs with human expertise, improving model robustness, and refining evaluation methodologies. By synthesizing trends and identifying research gaps, this study aims to guide future innovations in the application of LLMs for qualitative analysis.
\end{abstract}


\section{Introduction}
The presence of Artificial Intelligence (AI) has been intensifying in supporting various human activities. Generative AI, particularly Large Language Models (LLMs), has demonstrated significant potential for automating complex processes by processing and generating human-like text \cite{b1},\cite{b2},\cite{b3}. These models are trained on extensive textual data and designed to identify complex patterns in natural language, making them versatile tools for tasks such as machine translation, summarization, text editing, chatbot assistance, and even software code generation \cite{b4},\cite{b5}.

One of the most promising applications of LLMs is in qualitative analysis, a research approach focused on exploring and interpreting data to uncover patterns, themes, and categories. Traditional qualitative analysis methods, such as grounded theory or thematic analysis, often require intensive human effort and are prone to subjectivity \cite{b6},\cite{b8}. Moreover, these methods struggle with scalability, particularly when analyzing large datasets, such as hundreds of interviews or millions of social media posts. LLMs offer transformative solutions to these challenges, automating processes like open coding and theme extraction while providing consistent and reproducible outputs \cite{b14},\cite{b17}.

The integration of LLMs into qualitative research presents transformative opportunities and challenges, particularly in domains such as Software Engineering (SE). As highlighted by Bano \textit{et al.} \cite{b21}, LLMs have the potential to address longstanding challenges in qualitative research, such as the time-consuming nature of manual analysis, inconsistencies in interpretation by multiple simultaneous researchers, and scalability limitations. While LLMs can assist in automating tasks such as coding, theme identification, and pattern recognition, they also raise critical concerns related to contextual understanding, reproducibility, and ethical implications. These insights underscore the urgent need for systematic investigations into the use of LLMs in qualitative research.

The main contribution of this paper is to report the results of a systematic mapping aimed at analyzing the state of the art on the use of LLMs in qualitative analysis, with the goal of identifying trends, research gaps, and opportunities for future investigations. This study seeks to provide valuable insights for researchers looking to integrate LLMs into their qualitative analysis processes, guiding them toward works that offer detailed prompt data and information on their effectiveness. In doing so, the paper not only provides a comprehensive overview of the state of the art but also highlights the potential of these technologies to transform how large volumes of textual data are processed and interpreted.

This paper is structured as follows: Section \ref{sec:2} provides a brief background on qualitative analysis and LLMs. Section \ref{sec:3} describes the systematic mapping protocol and results. Section \ref{sec:4} presents a summary of contributions and research opportunities. Section \ref{sec:5} discusses the threats to validity. Section \ref{sec:6} presents the main conclusions. Finally, Section \ref{sec:7} presents the take away lessons.

\section{Background}
\label{sec:2}

\subsection{Limitations of Traditional Qualitative Methods}

While qualitative research methods, such as grounded theory or thematic analysis, provide valuable insights into complex phenomena, they face significant challenges. For instance, manual coding and categorization processes are notoriously labor-intensive, often requiring weeks or even months of effort to analyze large datasets effectively \cite{b6}. Additionally, these methods are prone to subjectivity, as researchers’ biases can unconsciously influence coding decisions, impacting the reliability and validity of findings \cite{b8}.

Moreover, traditional methods struggle with scalability. For example, analyzing hundreds of interviews or millions of social media posts in real time is impractical for human analysts. These limitations hinder the efficiency and breadth of qualitative analysis, particularly in rapidly evolving fields like software engineering or healthcare.

\subsection{Capabilities of LLMs in Overcoming Limitations}

LLMs offer transformative solutions to many challenges faced by traditional qualitative research methods. LLMs, such as GPT-4, LLaMA, and ChatGPT, are generative AI systems trained on vast datasets, enabling them to process and generate human-like text \cite{b4},\cite{b5}. By leveraging billions of parameters, these models excel in identifying patterns, performing thematic analysis, and automating coding processes.

\begin{itemize}

    \item Automation of Coding: LLMs can rapidly process and analyze text, performing open and axial coding in minutes. For example, ChatGPT has been employed to analyze qualitative survey responses, extracting key themes such as student satisfaction and resource accessibility in educational research \cite{b12},\cite{b11}.

\item Reducing Subjectivity: While human coding is prone to inconsistency, LLMs provide consistent and reproducible outputs, especially when fine-tuned or used with standardized prompts \cite{b14}. However, their adaptability and evolution over time may impact reproducibility. This minimizes bias and ensures greater reliability in qualitative findings, though care must be taken to account for potential variability in outputs across different versions of the models.

\item Scalability and Speed: Unlike human analysts, LLMs can handle vast amounts of unstructured data efficiently. For example, in healthcare studies, LLMs have been used to process patient feedback, identifying recurring themes in hours rather than weeks \cite{b17}.

\end{itemize}

\subsection{Illustrative Examples}

The following examples demonstrate the transformative potential of LLMs:

\begin{itemize}
    \item Healthcare Case Study: In a study analyzing patient feedback, an open-source LLM was fine-tuned to perform inductive thematic analysis, identifying critical concerns like delays in care and communication breakdowns \cite{b17}. This process significantly reduced the time required for analysis while maintainiong accuracy.

    \item Education Use Case: ChatGPT was used to categorize open-ended survey responses from students, identifying patterns such as dissatisfaction with online learning platforms and preferences for in-person classes \cite{b12}. The automated approach streamlined analysis, allowing researchers to focus on interpreting results.
    
\end{itemize}

\subsection{Motivation for this Study}

Despite their potential, LLMs face limitations that must be addressed for optimal application in qualitative research. For example, they are highly dependent on well-structured prompts, which, if poorly designed, can lead to inaccurate or incomplete outputs \cite{b19},\cite{b17}. Additionally, LLMs can generate ``hallucinations"—fabricated responses that lack basis in the data \cite{b13}. These challenges highlight the need for systematic studies to explore best practices, identify gaps, and refine methodologies for integrating LLMs into qualitative research.

This study aims to fill this gap by systematically mapping the current state of the art in LLM applications for qualitative research. By analyzing existing studies, this research identifies trends, challenges, and opportunities, providing a roadmap for researchers seeking to leverage LLMs effectively in their workflows.

The study by Leça \textit{et al.}\cite{b22} also converges in this direction by presenting a systematic mapping to investigate how LLMs are typically used in qualitative analysis and how they can be applied in Software Engineering research. Their study and this study complement each other in several aspects. However, while Leça \textit{et al.}\cite{b22} explore benefits (\textit{e.g.}, support in theme and pattern identification and flexibility for researchers) and limitations (\textit{e.g.}, ethical concerns and privacy), this study focuses specifically on the effectiveness of LLMs and the evaluation methods used, including specific metrics and criteria. Together, these studies provide a comprehensive perspective mapping the state of the art as a guide for future work and offering an analysis of the contextual implications and challenges in using these technologies. Both contribute to the understanding and applicability of LLMs in qualitative analysis.

\section{Systematic Mapping Study (SMS)}
\label{sec:3}
The structure adopted in this systematic mapping study was developed in accordance with the guidelines proposed by Kitchenham and Charters \cite{b10}. The main stages carried out include planning, conducting and reporting.

Parsif.al \footnote{\url{https://parsif.al/}} was used as a support tool for planning, conducting, and reporting the systematic mapping study, enabling the documentation and execution of the protocol.

\subsection{Planning}

The objective of this systematic mapping study was to identify the state of the art regarding the relationship between LLMs and qualitative analysis. Thus, the studies of interest include those that assess the application of LLMs in conducting qualitative analyses. Consequently, research questions and the corresponding protocol were established.
\begin{itemize}
    \item \textbf{RQ1}: What is the context in which LLMs have been applied to support qualitative analysis?
    
    \textbf{Rationale}: This question aims to identify the application contexts of LLMs, which is crucial for mapping the study areas where these models have demonstrated the most effectiveness.
    \item \textbf{RQ2}: What model and LLM configuration were used, and which data sources were utilized?
    
    \textbf{Rationale}: Focuses on the specific LLM models, their configurations, and the data sources used, given that different models and configurations can significantly influence the accuracy and effectiveness of qualitative analysis.
    \item \textbf{RQ3}: How was the LLM used to conduct qualitative analysis, and what techniques or methodologies were applied?
    
    \textbf{Rationale}: Investigates how LLMs were utilized in conducting qualitative analyses, including the techniques and methodologies applied. This question is essential for understanding best practices and adapting existing methodologies to maximize the potential of LLMs in different analysis contexts.
    \item \textbf{RQ4}: How was the effectiveness of the LLM evaluated, and what were the main outcomes?
    
    \textbf{Rationale}: Analyzes whether the effectiveness of LLMs was assessed and the key results obtained, which is essential for measuring the added value of this approach compared to traditional methods, ensuring the validation of its application in supporting qualitative research.
    \item \textbf{RQ5}: What limitations and future research directions were reported?
    
    \textbf{Rationale}: Examines the identified limitations and future research directions, contributing to the definition of continuous improvement strategies and refinement of the technology, ensuring that the use of LLMs becomes robust, accurate, and accessible in future applications.
\end{itemize}

\subsection{Search Strategy}

A generic search string was developed using the keywords ``LLM" and ``Qualitative Analysis". These keywords were connected using the logical operator AND, while their synonyms were linked with the OR operator. The terms in the search string were selected to ensure broader coverage of relevant studies. The string was tested in various configurations on the Scopus database, and after a calibration process, the final version defined was:

\begin{center}
(``LLM" \textbf{OR} ``LLMs" \textbf{OR} ``Large Language Model" \textbf{OR} ``Large Language Models") \textbf{AND} (``Qualitative Analysis" \textbf{OR} ``Qualitative Research" \textbf{OR} ``Grounded Theory")
\end{center}

In addition to Scopus\footnote{\url{http://www.scopus.com}}, searches using the search string were also conducted in the ACM Digital Library\footnote{\url{https://dl.acm.org/}}, IEEExplore\footnote{\url{https://ieeexplore.ieee.org/}}, Web of Science\footnote{\url{http://www.isiknowledge.com}}, and SBC Open Lib databases\footnote{\url{https://sol.sbc.org.br/busca/}}. The searches were carried out during the months of September and October 2024, using filters applied to titles, abstracts, and keywords. No restrictions were placed on the starting publication year. In total, 354 studies were retrieved.

Considering the emerging nature of the research topic and the likelihood of relevant studies not yet undergoing peer review, the Arxiv\footnote{\url{https://arxiv.org/}} database was also explored using keywords directly, without applying the search string. Arxiv papers were carefully considered due to their potential to provide early access to significant studies that are not yet indexed in major digital libraries like Scopus. Including these papers ensured a comprehensive exploration of the state of the art in LLMs for qualitative analysis.

\subsection{Selection Criteria}

Subsequently, selection criteria were established to support the proper identification of relevant studies for this systematic mapping. Titles, abstracts, and keywords were read and analyzed, with the selection criteria outlined in Table \ref{tab:criteria} (where IC refers to Inclusion Criteria, and EC refers to Exclusion Criteria).

It is important to note that EC1 is presented as the direct inversion of IC1 to emphasize the mutual exclusivity of these criteria, ensuring clarity in identifying studies that analyze the application of LLMs in qualitative analysis. The remaining exclusion criteria (EC2, EC3 and EC4) were defined independently to address specific attributes of the studies, such as language, accessibility, or type of publication.

\begin{table}[htbp]
\caption{Inclusion and exclusion criteria}
\begin{center}
\begin{tabular}{|c|p{6cm}|}
\hline
\rowcolor[HTML]{D3D3D3} \textbf{ID} & \textbf{Description} \\
\hline
\rowcolor{white} IC1 & The study analyzes the application of LLMs in qualitative analysis. \\
\hline
\rowcolor[HTML]{FFFFFF} EC1 & The study does not analyze the application of LLMs in qualitative analysis. \\
\hline
\rowcolor{white} EC2 & The study is not written in Portuguese or English. \\
\hline
\rowcolor[HTML]{FFFFFF} EC3 & The study is not openly accessible or available through institutional access. \\
\hline
\rowcolor{white} EC4 & The study is not a primary study. \\
\hline
\end{tabular}
\label{tab:criteria}
\end{center}
\end{table}
\FloatBarrier

\subsection{Data Extraction}

A questionnaire was developed to guide the data extraction process. This extraction criterion (DE, which refers to data extraction) is designed to address the research questions (RQ) defined in the study. The questions in the extraction form contain pre-defined response options to ensure consistency in data collection. Table \ref{tab:research_questions} presents the extraction criteria (DE) and maps each one to the corresponding research questions (RQ).

\begin{table}[htbp]
\caption{Data extraction and descriptors}
\begin{center}
\begin{tabular}{|p{0.5cm}|p{0.6cm}|p{3.5cm}|p{2.5cm}|}
\hline
\rowcolor[HTML]{D3D3D3} \textbf{RQ} & \textbf{DE} & \textbf{Description} & \textbf{Answer options} \\
\hline
\multirow{3}{*}{RQ1} 
& DE1* & Does the study primarily focus on analyzing the use of LLMs to support qualitative data analysis? & Yes; Partially; No \\
\cline{2-4}
& DE2 & What was the context of the qualitative analysis? & Education; Health; Culture; Law; Technology; Software Engineering; Others \\
\cline{2-4}
& DE3 & What was the main objective of the study in applying LLMs in qualitative analysis? & Compare with manual analysis; Automate analysis; Assess capabilities; Others \\
\hline
\multirow{3}{*}{RQ2} 
& DE4 & Which LLM model was used? & ChatGPT; Llama; NVivo; Atlas; Others \\
\cline{2-4}
& DE5 & Was the model adjusted or adapted for the study? & Yes; Partially; No \\
\cline{2-4}
& DE6 & What were the sources of data analyzed? & Interviews; Social media; Documents; Others \\
\hline
\multirow{4}{*}{RQ3} 
& DE7 & How was the qualitative analysis process conducted using the LLM? & Free text \\
\cline{2-4}
& DE8 & Were specific techniques/methodologies employed to support the analysis? & Yes; Partially; No \\
\cline{2-4}
& DE9 & What techniques were used or evaluated? & Open, axial, or deductive coding; Extraction of themes or topics; Categorization; Others \\
\cline{2-4}
& DE10 & Was any established qualitative analysis framework used? & Content analysis; Discourse analysis; Grounded Theory; Others \\
\hline
\multirow{4}{*}{RQ4} 
& DE11* & Does the article detail the prompt engineering performed? & Yes; No \\
\cline{2-4}
& DE12 & What metrics or criteria were used to evaluate the effectiveness of the LLM in qualitative analysis? & Comparison with human analysis; Accuracy; Precision; Recall; F1 Score; Others \\
\cline{2-4}
& DE13 & Did the study compare with traditional qualitative analysis methods? & Yes; No \\
\cline{2-4}
& DE14 & If the answer to question 12 is yes, was the LLM better, worse, or equal to traditional methods? & Better; Worse; Equal \\
\hline
\multirow{2}{*}{RQ5} 
& DE15 & Were specific limitations identified in the use of LLMs for qualitative analysis? & Yes; No \\
\cline{2-4}
& DE16 & Are there recommendations or suggestions from the authors for the use or future development of LLMs in qualitative analysis? & Yes; No \\
\hline
\end{tabular}
\label{tab:research_questions}
\end{center}
\end{table}
\FloatBarrier
The criteria (DE1 to DE16) were used to compose the prompt, and ChatGPT\footnote{\url{https://chatgpt.com/}} was employed in the data extraction stage of the studies analyzed in this systematic mapping. Polak and Morgan \cite{b11} reported results close to 90\% precision and recall for their proposed method, named ChatExtract, which leverages LLMs like GPT-4 in the data extraction process. Syriani, David, and Kumar \cite{b13} highlighted in their findings that the use of ChatGPT for automating the article screening process in systematic reviews shows promising potential, achieving an accuracy of 82\%.

Regarding data extraction DE7, open-ended responses were obtained, which involved manual content analysis, as well as consulting the full original text to clarify any potential doubts.

\subsection{Conducting}

\begin{figure}[htbp]
\centerline{\includegraphics[width=0.5\textwidth]{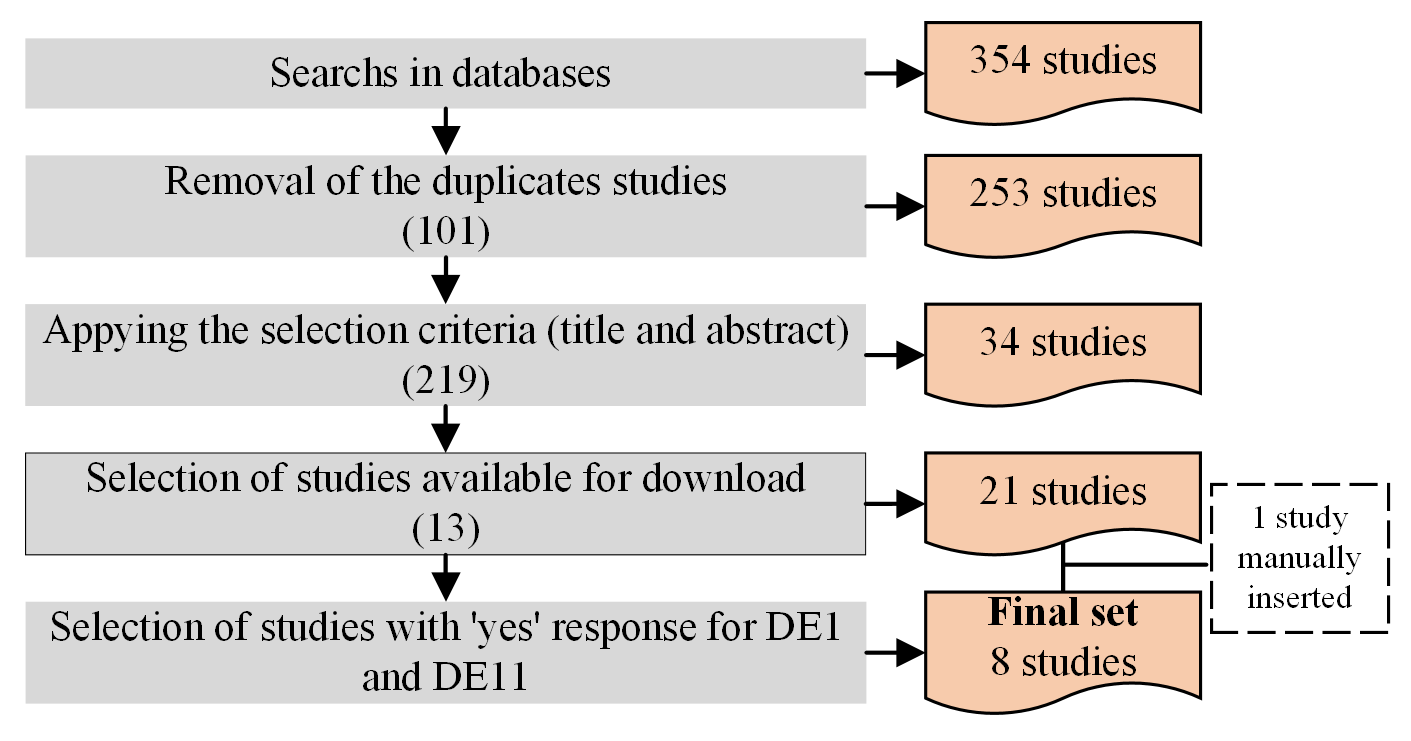}}
\caption{Study selection process.}
\label{fig1}
\end{figure}

Figure \ref{fig1} illustrates the steps followed in the study selection process, starting from the execution of the search string across the databases. A total of 354 studies were retrieved, distributed as follows: 20 from the ACM Digital Library, 30 from IEEExplore, 78 from Web of Science, 32 from SBC Open Lib, 193 from Scopus, and 1 from Arxiv.

Subsequently, a refinement process was conducted to select the final set of studies. First, duplicate studies were removed, resulting in 253 studies. Next, inclusion and exclusion criteria were applied based on the reading of titles and abstracts, reducing the total to 34 studies. In the following stage, only studies available for download were included, which further reduced the count to 21. 
The study \cite{b00} was a special case. It was retrieved but was excluded due to lack of full-text access. However, in January 2025, the authors published the complete version on ArXiv. We opted to manually include the study to complement the mapping.

If a study did not fully meet the response for criterion DE1 (i.e., the response was ``no" or ``partially"), the remaining criteria were not evaluated, and the study was excluded. Additionally, if a study did not meet criterion DE11 (``no"), it was excluded for not describing the prompt engineering used, which is essential to ensure reproducibility. 

At the end of this process, 8 studies remained, for which comprehensive data extraction was conducted. Table \ref{tab:selected_studies} presents all the studies that were included in this systematic mapping.

\begin{table}[htbp]
\caption{List of selected studies}
\begin{center}
\begin{tabular}{|p{0.3cm}|p{6.8cm}|p{0.4cm}|}
\hline
\rowcolor[HTML]{D3D3D3} \textbf{ID} & \textbf{Study Title} & \textbf{Ref.} \\
\hline
S1 & Artificial Intelligence and content analysis: the large language models (LLMs) and the automatized categorization & \cite{b13}\\ 
\hline
S2 & Coding Open-Ended Responses using Pseudo Response Generation by Large Language Models & \cite{b14}\\ 
\hline
S3 & Deep Learning Models for Analyzing Social Construction of Knowledge Online & \cite{b15}\\ 
\hline
S4 & Exploring Qualitative Research Using LLMs & \cite{b16}\\ 
\hline
S5 & Inductive thematic analysis of healthcare qualitative interviews using open-source large language models: How does it compare to traditional methods? & \cite{b17}\\ 
\hline
S6 & LLMusic: Topic Modeling in Song Lyrics Combining LLM, Prompt Engineering, and BERTopic & \cite{b18}\\ 
\hline
S7 & Performing an Inductive Thematic Analysis of Semi-Structured Interviews With a Large Language Model: An Exploration and Provocation on the Limits of the Approach & \cite{b19}\\
\hline
S8 & CollabCoder: A Lower-barrier, Rigorous Workflow for Inductive Collaborative Qualitative Analysis with Large Language Models & \cite{b00}\\
\hline
\end{tabular}
\label{tab:selected_studies}
\end{center}
\end{table}
\FloatBarrier

\subsection{Results}

The analysis revealed that 75\% of the included studies (6 out of 8) were published in 2024, with the remaining two in 2023. This reflects the recent emergence of LLMs in qualitative analysis and the growing interest of the scientific community in this field.

Regarding publication venues, most studies were concentrated in healthcare, education, and cultural studies, with no works identified in Software Engineering (SE). This gap is notable given the potential of LLMs to assist in SE tasks such as requirements engineering and user feedback analysis. A possible reason for this absence is the early-stage adoption of LLMs in SE. Future research could explore their integration in SE-specific venues, assessing their impact on traditionally manual processes.\\

\textbf{RQ1: What is the context in which LLMs have been applied to support qualitative analysis?}

Studies on the use of LLMs for qualitative analysis are concentrated in fields such as healthcare \cite{b14},\cite{b17}, education \cite{b13},\cite{b15},\cite{b19}, cultural studies \cite{b18}, and technological applications \cite{b16}.

Regarding the primary objectives, three studies evaluated the potential of LLMs to automate qualitative analysis \cite{b13},\cite{b15},\cite{b19}. The study \cite{b00} proposed CollabCoder, a workflow integrating LLMs into collaborative qualitative analysis, assisting in code suggestion, discussions, and codebook development. Other studies focused on fully automating qualitative data analysis \cite{b14},\cite{b18}, while \cite{b16},\cite{b17} compared LLM-assisted qualitative analysis with traditional methods.\\

\textbf{RQ2: What LLM model and configuration were used, and which data sources were utilized?}

The studies examined various LLM models, including ChatGPT \cite{b14},\cite{b16},\cite{b19},\cite{b00}, LLaMA-2 \cite{b17}, ATLAS.TI \cite{b13}, BERT \cite{b15} and Sabiá-2 medium \cite{b18}. About 50\% of the studies made adjustments or adaptations to the models for analysis \cite{b13},\cite{b14},\cite{b15},\cite{b17}. However, study \cite{b00} used ChatGPT without fine-tuning, relying solely on prompt engineering to assist in qualitative coding.

Regarding data sources, most studies analyzed interviews \cite{b13},\cite{b17},\cite{b19}, followed by documents \cite{b14},\cite{b15}, including \cite{b00}, which used book reviews as qualitative textual data. Other studies examined social media (app user reviews) \cite{b16} and song lyrics \cite{b18}.\\

\textbf{RQ3: How was the LLM used to conduct qualitative analysis, and what techniques or methodologies were applied?}

All included studies applied techniques and/or methodologies for qualitative analysis, also detailing the prompt engineering used. Theme extraction was combined with content analysis in two studies \cite{b13},\cite{b16}, while others integrated open coding and theme extraction \cite{b17},\cite{b19}. Two studies combined categorization and theme extraction, with one adopting the Grounded Theory approach \cite{b14} and the other using content analysis \cite{b15}. Study \cite{b00} followed an inductive approach, structuring open coding and theme extraction within a collaborative workflow based on Grounded Theory and Thematic Analysis. Study \cite{b18} tested three techniques: categorization, theme extraction, and topic extraction.  

LLM application methods varied among studies. Three employed prompt-based instruction \cite{b16},\cite{b17},\cite{b19}. Study \cite{b18} combined prompt engineering with automated extraction using BERT, while \cite{b15} exclusively used automated extraction with BERT. In \cite{b00}, ChatGPT provided automated code suggestions, facilitated coder discussions, and assisted in developing final code groupings within an AI-assisted workflow. Other studies applied automated extraction via ATLAS.TI, linked to specific techniques \cite{b13}, or conducted model training \cite{b14}.\\

\textbf{RQ4: How was the effectiveness of the LLM evaluated, and what were the main results?}

All seven included studies compared LLM-assisted qualitative analysis with traditional methods. In \cite{b13},\cite{b14},\cite{b15},\cite{b17},\cite{b19}, results were considered equivalent to traditional methods. Study \cite{b18} reported superior performance for LLMs, whereas \cite{b16} found LLM-assisted analysis to be less effective, particularly compared to human analysis. Study \cite{b00} assessed effectiveness through a user study comparing CollabCoder with Atlas.ti Web, revealing that CollabCoder improved efficiency by streamlining coder discussions and decision-making while maintaining high agreement rates. Participants also reported a lower learning curve and better workflow support compared to Atlas.ti Web.

The metrics used to evaluate performance included comparisons between LLM-assisted analysis and human analysis \cite{b13},\cite{b16},\cite{b17},\cite{b19}, comparisons with human analysis and accuracy \cite{b15}, and comparisons involving human analysis, accuracy, F1 Score, and precision \cite{b14},\cite{b18}. In \cite{b00}, the evaluation included metrics such as Cohen’s Kappa and Agreement Rate to assess coder consensus, demonstrating that the system facilitated alignment and reduced disagreement in the coding process.\\

\textbf{RQ5: What limitations and future research directions were reported?}

All studies identified specific limitations in using LLMs for qualitative analysis. One limitation mentioned was the reliance on well-structured prompts for accurate results \cite{b17},\cite{b19}. Another recurring limitation was the tendency of LLMs to generate ``hallucinations", producing responses without clear justification from the data \cite{b13},\cite{b19}. Additionally, inherent model biases were noted, particularly when dealing with sensitive information \cite{b17}. Other limitations included difficulties in assigning topics to subjective expressions \cite{b18}, lack of context sensitivity and emotional nuance \cite{b16}, and restrictions in interpreting complex meanings \cite{b19}. Study \cite{b14} also highlighted excessive response over-segmentation and over-categorization. In \cite{b00}, challenges were reported in achieving coder consensus when dealing with ambiguous data, requiring human intervention for nuanced interpretations.  

Further limitations included difficulties in broader and more detailed categorization, leading to a loss of specific details important by human researchers \cite{b13}, as well as inconsistencies in zero-shot classifications \cite{b18}. Another concern was the ``echo chamber" effect, with LLMs reproducing pre-trained patterns \cite{b16}. Study \cite{b00} also highlighted challenges in maintaining coder independence while fostering collaboration, along with risks of premature influence from LLM-generated coding suggestions that could bias human coders.

Regarding recommendations for LLM usage, studies \cite{b13},\cite{b17},\cite{b19} propose that LLMs serve as aids rather than replacements for human analysts, particularly in categorization, with researchers retaining final interpretative authority. Other works emphasize the importance of human collaboration and LLMs role in enhancing accuracy and validity \cite{b16},\cite{b19}. Study \cite{b00} suggests leveraging LLMs to structure and refine codebooks while ensuring human coders make final decisions. It also stresses the need for greater user control over AI-generated suggestions to prevent over-reliance on automated coding.  

For future research, there is broad agreement on the need to refine prompt engineering techniques and explore more advanced architectures to enhance LLMs' textual understanding \cite{b15},\cite{b16},\cite{b18},\cite{b19}. Increasing training data volume is also recommended to improve accuracy \cite{b15}, along with fine-tuning pre-trained models for better topic assignments \cite{b18}. Study \cite{b00} suggests optimizing prompt design to enhance AI-assisted code generation, refining interaction models for better coder collaboration and include integrating multimodal. Additional research directions include enhancing LLM capabilities to capture semantic and subjective nuances \cite{b13}. Study \cite{b17} suggestes that expanding open-source models for cost reduction and accessibility, and developing more user-friendly interfaces for researchers.\\

\section{Summary of Contributions and Research Opportunities}
\label{sec:4}

This section summarizes the main findings and contributions of this study, as well as the research opportunities identified. The key contributions include:

\textbf{Mapping of the field:}This study provides an overview of the use of LLMs in qualitative analysis, covering various aspects such as: (i) application contexts, (ii) LLM models and data sources, (iii) techniques and methodologies adopted, (iv) metrics used to evaluate effectiveness, and (v) limitations and research opportunities in the current state of the art.

This mapping focused on studies that reported LLM applications for qualitative analysis, detailing the prompt engineering employed—an essential factor for ensuring reproducibility. The findings highlight the recent nature of this field, reflected in the relatively low number of included studies, most of which were published in 2023 and 2024.

However, several research opportunities emerged. While LLMs exhibit performance comparable to or even superior to traditional methods in some cases, improvements are still needed. A key issue is their reliance on well-structured prompts, necessitating further exploration of approaches that enhance LLMs' textual understanding, allowing for greater flexibility in qualitative analyses.

Additionally, opportunities exist to investigate new architectures and techniques that improve the capture of semantic and subjective nuances, ensuring more precise and robust analyses. Another promising direction involves methods to mitigate ``hallucinations" and biases, such as fine-tuning models or employing validation and filtering techniques. The development of user-friendly interfaces for researchers without AI expertise is also a relevant area of exploration. Finally, establishing a standardized and robust evaluation metric capable of capturing the complexity of qualitative analysis performed by LLMs represents a significant avenue for future research.

\section{Threats to Validity}
\label{sec:5}

A concern when using LLMs to automate the selection of studies is the potential loss of knowledge that would normally be acquired during a full reading of the articles. Manual reviews allow researchers to better understand the field of study, identifying nuances and contexts that may go unnoticed in an automated analysis. This acquired knowledge can positively influence the quality of the analysis. To mitigate this risk, we conducted additional checks through comprehensive readings of the articles to clarify any doubts, ensuring a more solid and well-founded understanding.

Another concern is the dichotomy in using LLMs both for data extraction and for investigating their effectiveness in qualitative research, which could create a potential conflict of interest. This may lead to confirmation bias, where the use of the LLM influences results in a way that favors its own application. To ensure the objectivity and integrity of the study, we took rigorous measures, relying on solid references from the analyzed articles to ensure that conclusions were based on evidence and not merely on the inferences generated by the model.

Another limitation concerns the lack of combination with other techniques in the systematic mapping, such as snowballing. Including this technique could contribute to identifying additional relevant studies. According to Wohlin \textit{et al.} \cite{b20}, when snowballing is combined with database searches, the review shows a significant increase in the identification of primary studies, allowing gaps left by strategies based solely on database searches to be filled. The absence of this approach may have led to the oversight of relevant articles and represents an opportunity to improve the study.

Other factors that may have influenced the results presented in this systematic mapping include threats to data validity, as well as internal, external, and construct validity.

\textbf{Data Validity:} One concern related to data validity is that ChatGPT may exhibit a certain level of superficiality. While robust, it lacks the deep contextual understanding that a human researcher can achieve. This may result in information being extracted incompletely or without sufficient analytical depth. To mitigate this threat, we adopted a manual review approach. Whenever doubts arose regarding the accuracy of the information provided by the model, a full reading of the articles was conducted. This procedure aimed to ensure that critical information was not overlooked, thus providing a more accurate and contextualized analysis.

The decision to exclude studies that did not describe prompt engineering (DE11) reflects the critical need to ensure replicability in this systematic mapping. Given that the use of LLMs for qualitative analysis is a relatively new area of study, detailed descriptions of methodologies, including prompt engineering, are crucial for enabling comparisons and understanding results across studies. However, we recognize that this criterion may have restricted the number of included studies, particularly during the early stages of research in this field. Future iterations of this study could consider less restrictive criteria during the screening phase, enabling a broader analysis that includes studies with incomplete methodological details. This approach could provide valuable insights into the evolution of LLM applications for qualitative analysis.

\textbf{Internal Validity:} Since two researchers were involved in the article selection process, there is a risk that inclusion criteria might vary due to individual biases. Differences in interpretation could affect the consistency of study selection. To mitigate this threat, we established standardized selection criteria, requiring that articles specifically focus on the use of LLMs for qualitative analysis. In cases of uncertainty, the researchers held joint discussions to reach a consensus, ensuring greater uniformity in the selection process.

\textbf{External Validity:} Generalizing the results is crucial to ensure that the conclusions of this study can be applied to different contexts and diverse populations. To address this concern, studies were selected from various fields, including healthcare, education, technology, and others. This selection aimed to assess whether the techniques used and the results obtained would be replicable across multiple domains. Through this approach, we sought to demonstrate that the conclusions presented have the potential to be generalized beyond the initial scope of the study, thereby enhancing their applicability and relevance in various contexts.

\textbf{Construct Validity:} A critical issue involves whether the tools and methods used truly capture the concepts under investigation. To ensure that the prompts used were aligned with the research objectives, we adopted an iterative prompt engineering process in which different versions were tested and refined. These versions were made available on Zenodo\footnote{\url{https://doi.org/10.5281/zenodo.14177022}} to promote greater transparency. Additionally, test articles with known responses were used to validate whether the model’s outputs were consistent with expectations. This procedure provided greater confidence that ChatGPT correctly interpreted the questions and returned responses aligned with the study’s objectives.

\section{Conclusion}
\label{sec:6}
This paper presented a SMS on the use of LLMs in qualitative analyses, aiming to explore the state of the art in this area. Overall, the analysis of the studies revealed that most publications are recent (2023-2024), indicating that the field is still in its early stages but showing growing interest.

This study provides an overview of the field, addressing various dimensions such as contexts, LLM models, data sources, techniques and methodologies, metrics, limitations, and opportunities. The applications span areas like healthcare, education, culture, and technology, with objectives ranging from automating qualitative analysis to comparing results with traditional methods. Several models were utilized, with a predominance of adjustments made to LLMs to enhance outcomes. The effectiveness of LLMs was mostly evaluated as equivalent to traditional methods, though some limitations were identified, such as the dependence on prompts and the risk of biases.

As future work, we plan to develop a method using LLMs to automate the qualitative analysis process, particularly focusing on the coding stages, by exploring new architectures applied in different tools. For this new method, we also aim to incorporate improvements in prompt engineering techniques to achieve greater accuracy and robustness. The integration of LLMs into structured qualitative analysis workflows also presents an interesting direction for future research, as it may enhance coder agreement and decision-making in collaborative settings.

With continued advancements in this field, it is expected that LLMs may, in the future, play an even more significant role in qualitative analysis, becoming an indispensable tool for researchers dealing with large volumes of textual data.

\section{Take Away Lessons}
\label{sec:7}
The systematic mapping study provides valuable insights into the application of LLMs for qualitative analysis. Based on the evidence from the analyzed articles, several key takeaways emerge:

\subsection{Most Effective LLM According to the Evidence} 
ChatGPT and its variations (\textit{e.g.}, GPT-4) were the most frequently cited models in the included studies. These models demonstrated superior capabilities for tasks such as open coding and theme extraction when compared to other LLMs, such as LLaMA-2 or Atlas.TI \cite{b14},\cite{b16},\cite{b19}.

\subsection{Capabilities and Current Limitations}
\textbf{Open Coding}: LLMs like ChatGPT are effective in automating the open coding process, rapidly identifying themes and patterns in textual data \cite{b13},\cite{b14}.

\textbf{Axial Coding and Visualization}: Tasks like axial coding and generating graphical representations (\textit{e.g.}, mind maps and visual models in Atlas.TI) still require user intervention. LLMs cannot yet perform complex integrations or visualizations that demand contextual and nuanced decision-making.

\textbf{Efficiency Gains}: A key finding highlights LLMs' potential to drastically reduce qualitative coding time. For instance, Mathis \textit{et al.} \cite{b17} reported that LLMs shortened coding from weeks to hours, demonstrating their transformative impact on qualitative research workflows.

\subsection{Best Practices and Future Directions} 

While LLMs offer substantial benefits, their effective use requires structured prompts and careful oversight. Researchers should leverage LLMs for automation-friendly tasks, such as initial coding and theme identification, while reserving interpretative or creative processes for human analysts. Advances in LLM architectures and prompt engineering may further expand their role in complex stages of qualitative analysis.

\section*{Acknowledgment}
The authors thank ChatGPT 4.0 (paid version) for its support in data extraction for this systematic mapping. The model significantly accelerated the identification and selection of relevant articles. It was also used for text translation into English, followed by manual review to ensure accuracy and clarity. All translations and extractions were later verified by human reviewers for spelling, grammar, and data consistency. Thanks to CNPq (grant 312275/2023-4), FAPERJ (grant number E-26/204.256/2024), and Stone Co (funded research project 1006) for financial support.

\end{document}